\RequirePackage{fix-cm}
\documentclass[smallcondensed]{svjour3}     
\smartqed  
\usepackage{graphicx}

\usepackage{textcomp, amssymb, amsmath, color, subfigure, cite}

\definecolor{myGreen}{RGB}{0,150,0}
\definecolor{myBlue}{RGB}{50,100,200}
\definecolor{myRed}{RGB}{200,0,0}
\definecolor{myBrown}{RGB}{150,75,0}

\newcommand{\units}[1]{\; \text{#1}}
\newcommand{\e}[1]{\times 10^{#1}}

\graphicspath{ {figs/noise/}{figs/freq/}{figs/patlength/}{figs/expat/}{figs/exfreq/}{figs/errevo/}{figs/noise2/}{figs/lvsnoise/} }

\begin{document}

\title{Random pattern and frequency generation using a photonic reservoir computer with output feedback}

\titlerunning{Pattern and frequency generation using a photonic RC with output feedback}        

\author{Piotr Antonik \and Michiel Hermans \and Marc Haelterman \and Serge Massar}


\institute{P. Antonik, M. Hermans, S. Massar \at
           Laboratoire d'Information Quantique \\ 
           Universit\'{e} libre de Bruxelles \\
           Avenue F. D. Roosevelt 50, CP 224 \\
           Brussels, Belgium \\
           \email{pantonik@ulb.ac.be}           
           \and
           M. Haelterman \at
           Service OPERA-Photonique \\
           Universit\'{e} libre de Bruxelles, \\
           Avenue F. D. Roosevelt 50, CP 194/5 \\
           Brussels, Belgium
}

\date{}

\maketitle

\begin{abstract}
  Reservoir computing is a bio-inspired computing paradigm for processing time dependent signals. The performance of its analogue implementations matches other digital algorithms on a series of benchmark tasks. Their potential can be further increased by feeding the output signal back into the reservoir, which would allow to apply the algorithm to time series generation. This requires, in principle, implementing a sufficiently fast readout layer for real-time output computation. Here we achieve this with a digital output layer driven by a FPGA chip. We demonstrate the first opto-electronic reservoir computer with output feedback and test it on two examples of time series generation tasks: frequency and random pattern generation. We obtain very good results on the first task, similar to idealised numerical simulations. The performance on the second one, however, suffers from the experimental noise. We illustrate this point with a detailed investigation of the consequences of noise on the performance of a physical reservoir computer with output feedback. Our work thus opens new possible applications for analogue reservoir computing and brings new insights on the impact of noise on the output feedback.
\end{abstract}

\section{Introduction}

Reservoir Computing (RC) is a set of machine learning methods for designing and training artificial neural networks, introduced independently in \cite{jaeger2004harnessing} and in \cite{maass2002real}. The idea behind these techniques is that one can exploit the dynamics of a recurrent nonlinear network to process time series without training the network itself, but simply adding a general linear readout layer and only training the latter. This results in a system that is significantly easier to train (the learning is reduced to solving a system of linear equations \cite{lukovsevivcius2009survey}), yet powerful enough to match other algorithms on a series of benchmark tasks. RC has been successfully applied to, for instance, channel equalisation and chaotic series forecasting \cite{jaeger2004harnessing}, phoneme recognition \cite{triefenbach2010phoneme} and won an international competition on prediction of future evolution of financial time series \cite{NFC}. 

Reservoir computing allows to efficiently implement simplified recurrent neural networks in hardware, such as e.g. optical components. Optical computing has been investigated for decades as photons propagate faster than electrons, without generating heat or magnetic interference, and thus promise higher bandwidth than conventional computers \cite{arsenault2012optical}. RC would thus allow to build high-speed and energy efficient photonic devices. Several important steps have been taken towards this goal with electronic \cite{appeltant2011information}, opto-electronic \cite{paquot2012optoelectronic,larger2012photonic,martinenghi2012photonic}, all-optical \cite{duport2012all,brunner2012parallel,vinckier2015high} and integrated \cite{vandoorne2014experimental} experimental RC implementations reported since 2012.

The potential of these systems can be significantly increased by feeding the output signal back into the reservoir. It has been shown that this additional feedback allows the algorithm to solve long horizon prediction tasks, such as forecasting chaotic time series, which are impossible to solve otherwise \cite{jaeger2004harnessing}. It would also allow the setup to run autonomously, that is, produce an output signal without receiving any input signal, and thus make it capable of generating periodic time series \cite{antonik2016towards}. Implementing this idea experimentally requires, in principle, a sufficiently fast readout layer capable of generating and feeding back the output signal in real-time (such as, for instance, the analogue solutions proposed in \cite{smerieri2012analog,duport2016fully,antonik2017online}).

In this work we demonstrate the first photonic reservoir computer with such output feedback. The readout layer of the opto-electronic reservoir, based on previous experiments \cite{paquot2012optoelectronic,larger2012photonic}, is implemented on a FPGA chip, as in \cite{antonik2016online}. The use of high-speed dedicated electronics makes it possible to compute the output signal in real time, and thus feed it back into the reservoir. 
This results in a digital readout layer, that nevertheless allows one to investigate many of the issues that will affect a system with purely analogue feedback. The latter is a much more complicated experiment.
Indeed the only analogue output layers implemented so far on experimental reservoir computers were reported in \cite{smerieri2012analog,duport2016fully,vinckier2016autonomous}. Using them for output feedback would require adding an additional electronic circuit consisting of a sample and hold circuit, amplification, and multiplication  by the input mask. The present experiment allows one to investigate the benefits that output feedback has to offer to experimental reservoir computing, while anticipating the difficulties and limitations that will affect a fully analogue implementation. Such a two-step procedure, in which part of the experiment is analogue and part digital, is a natural procedure, and parallels the development of experimental reservoir computers in which some of the first experiments were only partially analogue, see e.g. \cite{appeltant2011information,martinenghi2012photonic}.

The system is tested on two time series generation tasks: sine wave with defined frequency and random pattern generation. We obtained very good results on the first task, matching perfectly previously reported numerical investigations \cite{antonik2016towards2}. Random pattern generation, however, being a more complex task, suffers from experimental imperfections -- namely, the noise -- that has not been taken into account previously \cite{antonik2016towards2}. As it markedly impacts the performance of the reservoir computer, we investigated, through numerical simulations, the effects of various levels of noise on an accurate model of the experimental setup. We found that by reducing the noise level the performance of the setup increases and ends up reaching the results obtained with an ideal, noiseless system. These simulations thus give an estimation of the results one could expect at a given level of noise.

This paper extends the preliminary investigation initiated in \cite{antonik2016pattern} by reporting an in-depth study of the effects of experimental noise on the performance of the reservoir computer, carried out through numerical simulations. Furthermore, this paper presents a broader overview of the experimental results outlined in \cite{antonik2016pattern}, and contains a more detailed description of the experimental setup and methods.

The manuscript is organised as follows. Sections \ref{sec:rc} and \ref{sec:tasks} introduce the concept of reservoir computing and the time series generation tasks, considered here. The experimental setup is described in section \ref{sec:exp}, and section \ref{sec:res} contains all the results, both experimental and numerical. Finally, section \ref{sec:ccl} concludes the paper.

\section{Reservoir computing with output feedback}
\label{sec:rc}

\begin{figure}
  \centering
  \includegraphics[width=0.8\textwidth]{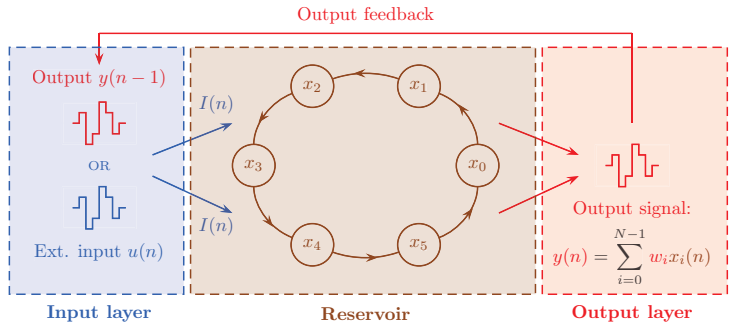}
  \caption{
    Conceptual representation of a reservoir computer with output feedback. The recurrent artificial neural network with $N$ nodes (in this scheme, $N=6$), denoted $x_i(n)$, has a ring-like topology. At training stage, the network is driven by an external time-multiplexed input signal $u(n)$ and has no output feedback. During autonomous run, the input signal $u(n)$ is switched off and the system is driven by its own output signal $y(n)$, given by a linear combination of the readout weights $w_i$ with the reservoir states $x_i(n)$.
  }
  \label{fig:rc}
\end{figure}

Fig. \ref{fig:rc} depicts a basic reservoir computer with output feedback. It contains a large number $N$ of internal variables $x_i(n)$ evolving in discrete time $n \in \mathbb{Z}$, as given by
\begin{equation}
  x_i(n+1) = f \left( \sum_{j=0}^{N-1} a_{ij} x_j(n) + b_i I(n) \right),
  \label{eq:rcevo}
\end{equation}
where $f$ is a nonlinear function, $I(n)$ is an input signal, injected into the system, and $a_{ij}$ and $b_i$ are time-independent coefficients, drawn from some random distribution with zero mean, that determine the dynamics of the reservoir. The variances of these distributions are adjusted to obtain the best performances on the task considered. 
All results, both experimental and numerical, presented in the present paper were obtained with a reservoir containing $N=100$ neurons.

The nonlinear function used here is $f = \sin (x)$, as in \cite{larger2012photonic, paquot2012optoelectronic}. To simplify the interconnection matrix $a_{ij}$, we exploit the ring topology, proposed in \cite{rodan2011minimum,appeltant2011information}, so that only the first neighbour nodes are connected. This architecture provides performances comparable to those obtained with complex interconnection matrices, as demonstrated numerically in \cite{lukovsevivcius2009survey,rodan2011minimum} and experimentally in \cite{appeltant2011information,larger2012photonic,paquot2012optoelectronic,duport2012all,brunner2012parallel}. Under these conditions Eq. \eqref{eq:rcevo} becomes
\begin{subequations}
  \begin{align}
    x_0(n+1) & = \sin \left(  \alpha x_{N-1}(n-1) + \beta M_0 I(n) \right),\label{eq:rcevo2_1} \\
    x_i(n+1) & = \sin \left(  \alpha x_{i-1}(n) + \beta M_i I(n) \right),\label{eq:rcevo2_2}
  \end{align}%
  \label{eq:rcevo2}%
\end{subequations}
with $i=1,\ldots,N-1$, $\alpha$ and $\beta$ parameters are used to adjust the feedback and the input signals, respectively, and $M_i$ is the input mask, drawn from a uniform distribution over the the interval $[-1, +1]$, as in \cite{rodan2011minimum, paquot2012optoelectronic, duport2012all}. 

The reservoir computer produces an output signal $y(n)$, given by a linear combination of the states of its internal variables
\begin{equation}
  y(n) = \sum_{i=0}^{N-1} w_i x_i (n),
  \label{eq:rcout}
\end{equation}
where $w_i$ are the readout weights, trained either offline (using standard linear regression methods, such as the ridge regression algorithm \cite{tikhonov1995numerical} used here), or online \cite{antonik2016online}, in order to minimise the Normalised Mean Square Error (NMSE) between the output signal $y(n)$ and the target signal $d(n)$, given by
\begin{equation}
  \text{NMSE} = \frac{ \left\langle \left( y(n) - d(n) \right)^2 \right\rangle }%
                     { \left\langle \left( d(n) - \langle d(n) \rangle \right)^2 \right\rangle}.
  \label{eq:mse}
\end{equation}

The input signal $I(n)$ can be either an external signal $I(n) = u(n)$, or the reservoir's own output, delayed by one timestep $I(n) = y(n-1)$. The system is operated in two stages: a training phase and an autonomous run. During the training phase, the reservoir computer is driven by a time-multiplexed teacher signal $I(n) = u(n)$, and the resulting states of the internal variables $x_i(n)$ are recorded. The teacher signal depends on the task under investigation (which will be introduced in section \ref{sec:tasks}). The system is trained to predict the next value of the teacher time series from the current one, that is, the readout weights $w_i$ are optimised so as to get as close as possible to $y(n) = u(n+1)$.
Then, the reservoir input is switched from the teacher sequence to the reservoir output signal $I(n) = y(n-1)$, and the system is left running autonomously. The reservoir output $y(n)$ is used to to evaluate the performance of the experiment.

\section{Time series generation tasks}
\label{sec:tasks}

Feeding the output back into the reservoir allows the computer to autonomously (i.e. without any external input) generate time series. We tested the capacity of the experiment to generate two examples of periodic signals: sine waves with defined frequencies \cite{wyffels2014frequency,antonik2016towards2,jaeger2007echo} and random patterns of various lengths \cite{wyffels2008stable,caluwaerts2013locomotion,reinhart2012regularization}.

\subsection{Frequency generation}
\label{subsec:taskfreq}
Frequency generation is the simplest time series generation task considered here. The system is trained to generate a sine wave given by
\begin{equation}
  u(n) = \sin\left( \nu n \right),
  \label{eq:freq}
\end{equation}
where $\nu$ is a real-valued relative frequency. The physical frequency $f$ of the sine wave depends on the experimental roundtrip time $T$ (see section \ref{sec:exp}) as follows
\begin{equation}
  f = \frac{\nu}{2\pi T}.
  \label{eq:freq2}
\end{equation}
This task allows to measure the bandwidth of the system and investigate different timescales within the neural network.

As the generated frequency is never exactly the same as desired, the resulting phase accumulation shifts the output signal from the target, thus rendering point-by-point error estimation inappropriate. For this reason we used the FFT algorithm to compute the frequency of the reservoir output signal and compare it to the frequency of the target signal \cite{antonik2016towards}.
If the frequency difference (error) is smaller than a certain threshold -- that we have set, arbitrarily, to $10^{-3}$ -- the generated frequency is considered as correct.

\subsection{Random pattern generation}
\label{subsec:taskpat}
Random pattern generation is a natural step forward from the frequency generation task to a more complex problem -- instead of a regularly-shaped continuous function, the system is trained to generate a periodic, but arbitrarily-shaped discontinuous function. Specifically, a pattern is a short sequence of $L$ randomly chosen real numbers (here drawn from the uniform distribution over the interval $\left[-0.5, 0.5 \right]$) that is repeated periodically to form an infinite time series \cite{antonik2016towards}.
Similarly to the physical frequency in section \ref{subsec:taskfreq}, the physical period of the pattern is given by $\tau_\text{pattern} = L \cdot T$.
The aim is to obtain a stable pattern generator, that reproduces precisely the pattern and does not deviate to another periodic behaviour. To evaluate the performance of the generator, we compute the NMSE between the reservoir output signal and the target pattern signal during the training phase and the autonomous run.

For some of our analysis we need to establish a criteria to decide whether a pattern is well reproduced or not. Similar to frequency generation (see section \ref{subsec:taskfreq}), we have decided -- again, arbitrarily -- to set this threshold at $10^{-3}$. That is, a pattern generated with a $\text{NMSE}<10^{-3}$ is considered satisfactory, while if $\text{NMSE}>10^{-3}$ the generated pattern is considered unsatisfactory. Of course, other thresholds are possible, and in the case of a practical application it would need to be set by the application.

\section{Experimental setup}
\label{sec:exp}

\begin{figure}
  \centering
  \includegraphics[width=0.7\textwidth]{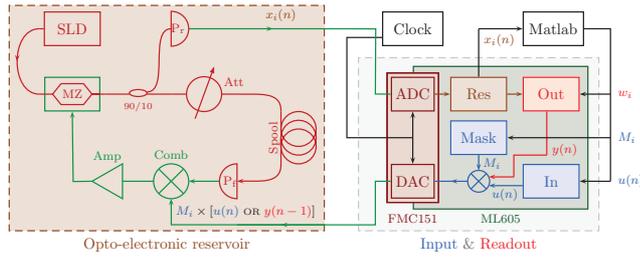}
  \caption{\textbf{(a)} Schematic representation of the experimental setup. Optical and electronic components of the photonic reservoir are shown in red and green, respectively. It contains an incoherent light source (SLD), a Mach-Zehnder intensity modulator (MZ), a $90/10$ beam splitter, an optical attenuator (Att), a fibre spool (Spool), two photodiodes ($\text{P}_\text{r}$ and $\text{P}_\text{f}$), a resistive combiner (Comb) and an amplifier (Amp). The FPGA board implements the readout layer and computes the output signal $y(n)$ in real time. It also generates the analogue input signal $I(n)$ and acquires the reservoir states $x_i(n)$.
  The computer, running Matlab, controls the devices, performs the offline training and uploads all the data ($u(n)$, $w_i$ and $M_i$) on the FPGA.
  }
  \label{fig:exp}
\end{figure}

Our experimental setup, schematised in Fig. \ref{fig:exp}, consists of two main components: the opto-electronic reservoir and the FPGA board.

\subsection{Opto-electronic reservoir}
\label{subsec:oeres}

The opto-electronic reservoir is based on previously published works \cite{paquot2012optoelectronic,larger2012photonic}. The reservoir states are encoded into the intensity of incoherent light signal, produced by a superluminiscent diode (Thorlabs SLD1550P-A40). The Mach-Zehnder (MZ) intensity modulator (EOSPACE AX-2X2-0MSS-12) implements the nonlinear function, its operating point is adjusted by applying a bias voltage, produced by a Hameg HMP4040 power supply.  A fraction (10\%) of the signal is extracted from the loop and sent to the readout photodiode (TTI TIA-525I) and the resulting voltage signal is sent to the FPGA. An optical attenuator (Agilent 81571A) is used to set the feedback gain $\alpha$ of the system (see Eqs. \eqref{eq:rcevo2_1} and \eqref{eq:rcevo2_2}). 
The fibre spool consists of approximately $1.6 \units{km}$ single mode fibre, giving a round trip time of $7.93 \units{\textmu s}$.
The resistive combiner sums the electrical feedback signal, produced by the feedback photodiode (TTI TIA-525I), with the input signal from the FPGA to drive the MZ modulator, with an additional amplification stage of $+27 \units{dB}$ (ZHL-32A+ coaxial amplifier) to span the entire $V_\pi$ interval of the modulator. 
The $N=100$ reservoir states $x_i$ are sampled at $203.7832\units{MHz}$ and averaged over 16 samples in order to get rid of the noise and the transients, induced by the finite bandwidth of the Digital-to-Analogue Converter (DAC).

\subsection{FPGA board}
\label{subsec:fpga}

In this work we use a Xilinx ML605 evaluation board, powered by a Virtex 6 XC6VLX240T FPGA chip. The board is paired with a 4DSP FMC151 daughter card, containing one two-channel Analogue-to-Digital Converter (ADC) and one two-channel DAC. The ADC's maximum sampling frequency is $250 \units{MHz}$ with 14-bit resolution, while the DAC can sample at up to $800 \units{MHz}$ with 16-bit precision. 

The FPGA board is used to interface the opto-electronic reservoir with a personal computer, running Matlab. The communication is operated via a fast Gbit Ethernet connection. First, the input mask $M_i$ and the teacher signal $u(n)$, generated in Matlab, are uploaded on the board, which then generates the masked input signal $M_i \times u(n)$, sent to the reservoir via the DAC. The resulting reservoir states $x_i(n)$ are sampled and averaged by the FPGA, and then sent to the computer in real time. That is, the design allows to capture the reservoir states for any desired time interval. After training of the reservoir using ridge regression algorithm \cite{tikhonov1995numerical}, the optimal readout weights $w_i$ are uploaded on the board.
Because of the relatively long delay needed for the offline training, the reservoir needs to be reinitialised in order to restore the desired dynamics of the internal states prior to running it autonomously. For this reason, we drive the system with an initialisation sequence of $128$ timesteps, before coupling the output signal with the input and letting the reservoir computer run autonomously. 
In this stage, the FPGA computes the output signal $y(n)$ in real time, then creates a masked version $M_i \times y(n)$ and sends it to the reservoir via the DAC.

As the neurons are processed sequentially, due to propagation delay between the intensity modulator (MZ) and the ADC, the output signal $y(n-1)$ can only be computed in time to update the 24-th neuron $x_{23} (n)$. For this reason, we set the first 23 elements of the input mask $M_i$ to zero. That way, all neurons contribute to solving the task, but the first 23 do not ``see'' the input signal.

The arithmetic operations computed by the FPGA are performed on real numbers. However, the chip is a logic device, designed to operate bits. The performance of the design thus highly depends on the bit-representation of real numbers, i.e. the precision. The main constraint comes from the ADC and DAC, limited to 14 and 16 bits, respectively. Numerical simulations, reported in \cite{antonik2016towards}, show that such precision is sufficient for this application. It was also shown in \cite{antonik2016towards} that the precision of the readout weights $w_i$ has a significant impact on the performance of the system. For this reason we designed the experiment for optimal utilisation of the resolution available. The reservoir states were tuned to lie within a $]-1,+1[$ interval. They are thus represented as 16-bit integers, with 1 bit for the sign and 15 bits for the decimal part. Another limitation comes from DSP48E slices, used to multiply the states $x_i(n)$ by the readout weights $w_i$. These blocks are designed to multiply a 25-bit integer by a 18-bit integer. To meet these requirements, we represent the $w_i$ as 25-bit integers, with 1 sign bit and 24 decimal bits. In order to keep the readout weights within the $]-1,1[$ interval, we amplify the reservoir states digitally inside the FPGA. That is, the $x_i(n)$ are multiplied by 8 after acquisition, prior to computing the output signal $y(n)$.

\section{Results}
\label{sec:res}

In this section, we present the experimental results obtained for the two tasks outlined in section \ref{sec:tasks}. We then give an in-depth analysis of the experimental noise, encountered in this work, and show how it affects the performance of the reservoir computer.

\subsection{Frequency generation}

For this task, our experimental results matched accurately the numerical predictions reported in \cite{antonik2016towards2}. Concretely, we expected a bandwidth of $\nu \in \left[ 0.06, \pi \right]$ with a 100-neuron reservoir.
The upper limit is a signal oscillating between $-1$ and $1$ and is given by half of the sampling rate of the system (the Nyquist frequency). The lower limit is caused by the memory limitation of the reservoir. In fact, low-frequency oscillations correspond to longer periods, and the neural network can no longer ``remember'' a sufficiently long segment of the sine wave so as to keep generating a sinusoidal output.
These numerical results are confirmed experimentally in the present work.

\begin{figure}
  \centering
    \includegraphics[width=0.7\textwidth]{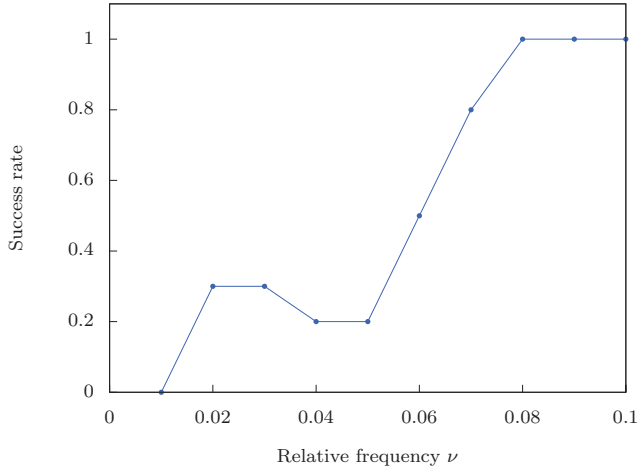}
    \caption{Determination of the lower limit of the reservoir computer bandwidth. Frequencies above $0.08$ are generated perfectly with any of the 10 random input mask, and are therefore not shown on the plot. Frequencies below $0.05$ fail with most input masks. We thus consider $0.08$ as the lower limit of the bandwidth, but keep in mind that frequencies as low as $0.02$ could also be generated, but only with a carefully picked input mask.
}
  \label{fig:bw}
\end{figure}

We tested our setup on frequencies ranging from $0.01$ to $\pi$, and found that frequencies within $[0.1, \pi]$ are generated accurately with any random input mask. Lower frequencies between $0.01$ and $0.1$, however, were produced properly with some random masks, but not all. For this reason, we investigated the $[0.01, 0.1]$ interval more precisely, since this is where the lower limit of the bandwidth lies. For each frequency, we ran the experiment 10 times for 10k timesteps with different random input masks and counted the number of times the reservoir produced a sine wave with the desired frequency and constant amplitude of $1$. The results are shown in Fig. \ref{fig:bw}. Frequencies below 0.05 are not generated correctly with most input masks. At $\nu=0.07$ the output is correct most of the times, and for $\nu=0.08$ and above the output sine wave is correct with any input mask. The bandwidth of this experimental RC is thus $\nu \in \left[ 0.08, \pi \right]$. 
Given the roundtrip time $T=7.93\units{\textmu s}$, this results in a physical bandwidth of $1.5$ -- $63\units{kHz}$.
Note that frequencies within this interval can be generated with any random input mask $M_i$. Lower frequencies, down to $0.02$, could also be generated, but only with a suitable input mask.

\begin{figure}
  \centering
  \includegraphics[width=0.7\textwidth]{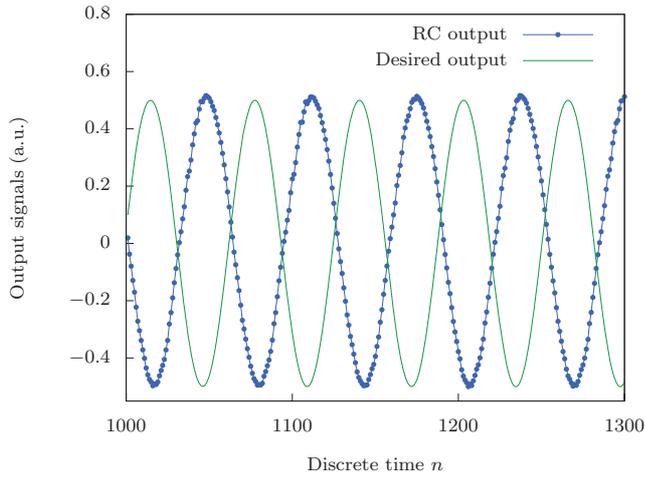}
  \caption{Example of an autonomous run output signal for frequency generation task with $\nu=0.1$. The experiment continues beyond the range of the figure.}
  \label{fig:exfreq}
\end{figure}

Fig. \ref{fig:exfreq} shows an example of the output signal during the autonomous run. The system was trained for $1000$ timesteps to generate a frequency of $\nu = 0.1$, and successfully accomplished this task with a frequency error of $7.5\e{-5}$.

\subsection{Random pattern generation}
\label{subsec:respat}

\begin{figure}[t]
  \centering
  \subfigure[]{\includegraphics[width=0.45\textwidth]{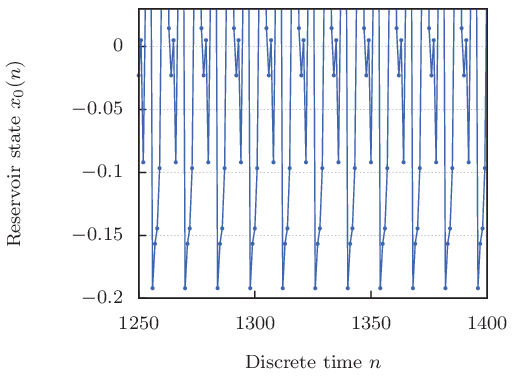}\label{subfig:cleannrn}}
  \subfigure[]{\includegraphics[width=0.45\textwidth]{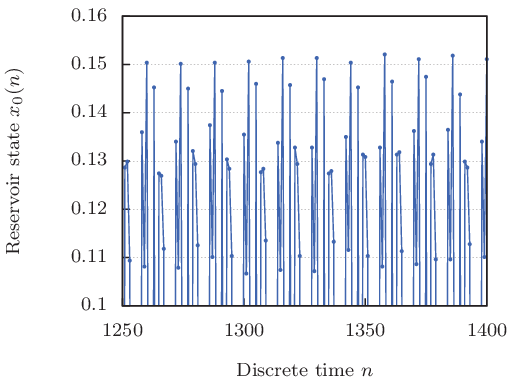}\label{subfig:noisynrn}}
  \caption{Examples of neuron behaviour in a \textbf{(a)} noiseless numerical and a \textbf{(b)} noisy experimental reservoir. For clarity, the range of the $y$-axis is limited to the area of interest. In the former case $x_i(n)$ cycles between several identical values, while in the latter it takes many similar, but not identical values.
  }
  \label{fig:noise1}
\end{figure}

The random pattern generation task is more complex than frequency generation and thus more sensitive to experimental imperfections. As will be discussed more in detail in section \ref{subsec:resnoise}, this task is markedly affected by the experimental noise. The goal of this task is two-fold: to ``remember'' a pattern of a given length $L$ and be able to reproduce it for an unlimited duration. We have shown numerically that a noiseless 51-neuron reservoir is capable of generating patterns up to 51-element long \cite{antonik2016towards}. This is a logical result, as, intuitively, each neuron of the system is expected to ``memorise'' one value of the pattern. Similar conclusions have been drawn in an earlier study of the memory capacity in echo state networks \cite{jaeger2001short}. Simulations of a noisy 100-neuron reservoir, similar to the experimental setup, show that the maximum pattern length is reduced down to $L=13$. This means that noise significantly reduces the effective memory of the system. In fact, the noisy neural network has to take into account the slight deviations of the output from the target pattern so as to be able to follow the pattern disregarding these imperfections. Fig. \ref{fig:noise1} illustrates this issue. The left panel depicts the behaviour of one neuron in a noiseless reservoir, driven by a periodic teacher signal. The reservoir state cycles between several identical values. The right panel shows what happens in our experimental system. While the noisy neuron exhibits periodic behaviour, it cycles between many similar, but not identical values.
This makes the generation task much more complex, and requires more memory, hence the maximal pattern length is shorter. 

\begin{figure}
  \centering
  \includegraphics[width=0.7\textwidth]{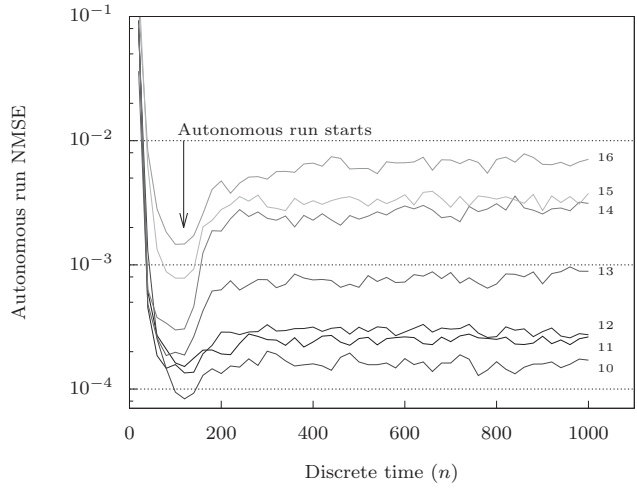}
    \caption{%
     Evolution of NMSE during experimental autonomous generation of periodic random patterns of lengths $L=10, \ldots, 16$ as function of discrete time $n$. The autonomous run starts at $n=128$, as indicated by the arrow. Patterns shorter than 13 are reproduced with low $\text{NMSE}<10^{-3}$, while patterns longer than 14 are not generated correctly with $\text{NMSE}>10^{-3}$. In the latter cases, the reservoir dynamics remains stable and periodic, but the output only remotely resembles the target pattern.}
  \label{fig:errevo}
\end{figure}

These predictions were also confirmed experimentally. Fig. \ref{fig:errevo} shows the evolution of the NMSE measured during the first 1k timesteps of a 10k-timestep autonomous runs with different pattern lengths. Plotted curves are averaged over 100 runs of the experiment, with 5 random input masks and 20 random patterns for each length $L$. The initial minimum (at $n=128$) corresponds to the initialisation of the reservoir (see section \ref{subsec:fpga}), then the output is coupled back and the system runs autonomously. Patterns with $L=12$ or less are generated very well and the error stays low. Patterns of length 13 show an increase in NMSE, but they are still generated reasonably well. For longer patterns, the system deviates to a different periodic behaviour, and the error grows above our $10^{-3}$ threshold (see section 3.2 for a discussion of the threshold).

\begin{figure}
  \centering
  \includegraphics[width=0.7\textwidth]{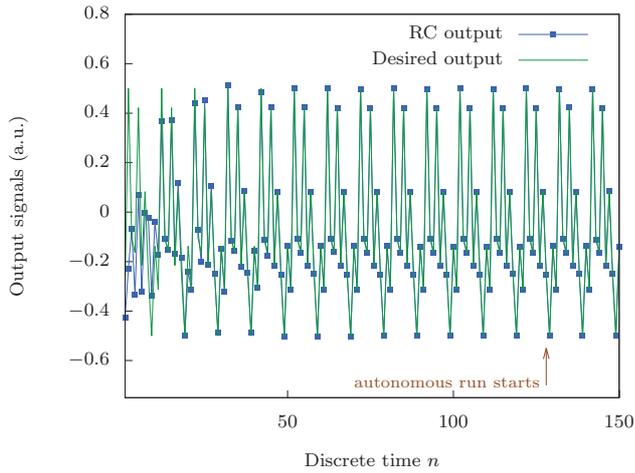}
  \caption{Example of an output signal for random pattern generation task, with a pattern of length 10. The reservoir is first driven by the desired signal for $128$ timesteps (see section \ref{subsec:fpga}), and then the input is connected to the output. Note that in this example the reservoir output requires about 50 timesteps to match the driver signal. The autonomous run continues beyond the scope of the figure.}
  \label{fig:expat}
\end{figure}

Fig. \ref{fig:expat} shows an example of the output signal during the autonomous run. The system was trained for $1000$ timesteps to generate a pattern of length 10. The reservoir computer successfully learned the desired pattern and the output perfectly matches the target signal. 

We also tested the stability of the generator by running it for several hours ($\sim10^9$ timesteps) with random patterns of lengths 10, 11 and 12. The output signal was visualised on a scope and remained stable and accurate through the whole test.

\subsection{Numerical study of the impact of noise}
\label{subsec:resnoise}

As have been mentioned in section \ref{subsec:respat}, the experimental noise has a significant impact on the performance of the reservoir computer. In fact, previously reported simulations \cite{antonik2016towards} considered an ideal noiseless reservoir, while our experiment is noisy. 
This noise is generated by the active components of the setup: the amplifier, which has a relatively high gain and is therefore very sensitive to small parasitic signals on the input, the DAC and the photodiodes. 
We found that each component of the setup contributes more or less equally to the overall noise level. Thus, it can not be reduced by replacing one ``faulty'' component. Neither can it be averaged out, as the output value has to be computed at each timestep.

\begin{figure}
  \centering
  \includegraphics[width=0.70\textwidth]{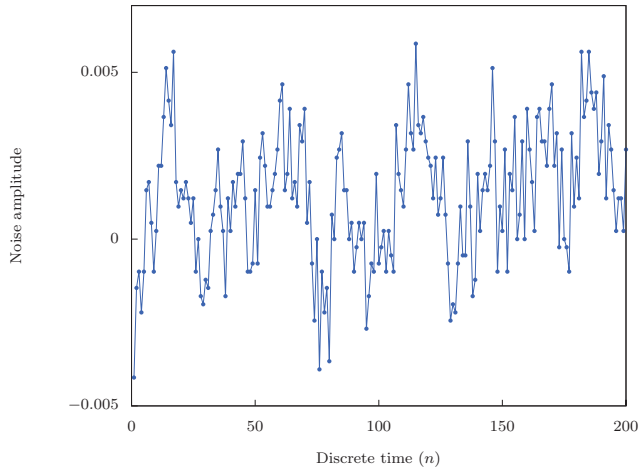}
  \caption{
    Illustration of the noise inside the experimental reservoir. 
    Time-multiplexed experimental reservoir states $x_i (n)$ are shown, that is, the first $100$ values correspond to $x_{0 \ldots 99} (0)$ and the next $100$ values are $x_{0 \ldots 99} (1)$.  The sequence was recorded in the case when the input signal is null $I(n)=0$, scaled so that in normal experimental conditions (non-zero input) the neurons would lie in a $[-1,1]$ interval. 
  }
  \label{fig:noise2}
\end{figure}

We estimated the level of noise present in the experiment as follows. We recorded the time-multiplexed reservoir states, received by the readout photodiode, as depicted in Fig. \ref{fig:noise2}. The system did not receive any input signal $I(n) = 0$. We used the recorded sequence to evaluate the noise level by taking the standard deviation of the signal, which gives $2\e{-3}$. For reference, the standard deviation of the time-multiplexed reservoir states under normal experimental conditions, with non-zero input, is roughly $0.2$.
The reason why we use standard deviation, and not the more common Signal-to-Noise Ratio, is because the standard deviation is more suitable to generate similar levels of noise in numerical simulations, as will be explained in what follows.

Since the noise plays such an important role, we performed a series of numerical experiments with different levels of noise to find out to what extent if affects the performance of the computer. We used a precise model of the experiment, based on previous works \cite{paquot2012optoelectronic,antonik2016towards},  that emulates the most influential features of the experimental setup, such as the high-pass filter of the amplifier, the finite resolution of the ADC and DAC, and accurate input and feedback gains. We added Gaussian white noise with zero mean and standard deviations ranging from $10^{-2}$ to as low as $10^{-8}$. These simulations allow to estimate the expected performance of the experiment for different levels of noise.

\begin{figure}
  \centering
  \includegraphics[width=0.7\textwidth]{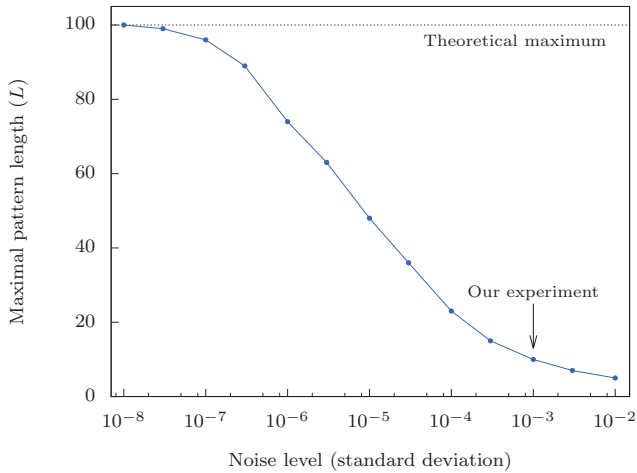}
  \caption{Impact of experimental noise on the performance of a reservoir computer with output feedback. The graph presents numerical results obtained with an accurate model of the experimental setup. Noise levels are shown as standard deviations of the Gaussian noise used in the simulations. The system was tested on the random pattern generation task and the performance metric is the maximal length $L$ of a pattern that the reservoir could generate. The theoretical maximum is $L=100$, since we used a reservoir with $N=100$ neurons. Noise levels of $10^{-8}$ and below are equivalent to an ideal noiseless system. The arrow indicates the experimental results presented in this work (see section \ref{subsec:respat}).}
  \label{fig:lvsnoise}
\end{figure}

Fig. \ref{fig:lvsnoise} shows the maximum pattern length $L$ that the reservoir computer is able to generate for different levels of noise. 
The maximal length is determined using the $10^{-3}$ autonomous error threshold, as described in section \ref{subsec:respat}. That is, if the NMSE does not grow above $10^{-3}$ during the autonomous run, the reservoir computer is considered to have successfully generated the given pattern.
For statistical purposes, we used 10 different random patterns for each length $L$ (and only counted the cases where the system have succeeded in all 10 trials).
These results show that the noise level of $10^{-8}$ is equivalent to an ideal noiseless reservoir. As the noise level increases, the memory capacity of the reservoir deteriorates. At a level of $10^{-3}$, the maximum pattern length is decreased down to 10, which matches the experimental results discussed in section \ref{subsec:respat}. For higher noise levels the results are, obviously, even worse. 

Overall, these results show what level of noise one should aim for in order to obtain a certain performance from an experimental reservoir computer with output feedback. Our experiments have confirmed the numerical results for the noise level of $10^{-3}$. In principle, one could double the maximal pattern length by carefully re-building the same experiment with low-noise components, namely a weaker amplifier and a low-$V_\pi$ intensity modulator, which would lower the noise to $10^{-4}$. Switching to a passive setup, such as the coherently driven cavity reported in \cite{vinckier2015high}, could potentially lower the noise down to $10^{-5}$ or even $10^{-6}$, with performance approaching the maximum memory capacity.

\section{Conclusion}
\label{sec:ccl}

In the present work, we reported the first photonic reservoir computer with output feedback, realised with a digital FPGA-based output layer. The experiment was able to successfully solve two time series generation tasks. We have also investigated the impact of experimental noise on systems with output feedback, and how it affects the performance.

Photonic implementations of recurrent artificial neural networks with output feedback can find many interesting applications. Such high-speed devices could be applied to signal generation in telecommunication applications, or high-frequency trading \cite{aldridge2009high}, with high requirements on the execution speed. Furthermore, precise pattern generation is required for efficient robot control. Adding the online training \cite{antonik2016online} makes the robot potentially capable of learning to move and adapting its motion to variable conditions. A reservoir computer could also be used as a function generator with tunable frequency \cite{wyffels2014frequency} or even store several different patterns and make use of the input signal to select the pattern to generate \cite{sussillo2009generating}. This work is thus the first step towards numerous additional applications of photonic reservoir computing.

\section*{Compliance with Ethical Standards}
\textbf{Funding.} This study was funded by the Interuniversity Attraction Poles program of the Belgian Science Policy Office (grant IAP P7-35 ``photonics@be''), by the Fonds de la Recherche Scientifique FRS-FNRS and by the Action de Recherche Concert\'{e}e of the Acad\'{e}mie Universitaire Wallonie-Bruxelles (grant AUWB-2012-12/17-ULB9).

\noindent
\textbf{Conflict of Interest.} All authors declare that they have no conflict of interest.

\noindent
\textbf{Ethical approval.} This article does not contain any studies with human participants or animals performed by any of the authors.



\end{document}